\documentclass{article}
\usepackage{spconf,amsmath,graphicx}
\usepackage{multirow}
\usepackage{url}

\title{Iterative learning for instance segmentation}
\name{Tuomas Sormunen$^{\star }$ Arttu L\"{a}ms\"{a}$^{\star }$, Miguel Bordallo López$^{\star\dagger}$}

\address{$^{\star}$ VTT Technical Research Centre of Finland\\
        $^{\dagger}$ Center for Machine Vision and Signal Analysis (CMVS), University of Oulu, Finland }

\begin{document}
%
\maketitle

%
\begin{abstract}
Instance segmentation is a computer vision task where separate objects in an image are detected and segmented. State-of-the-art deep neural network models require large amounts of labeled data in order to perform well in this task. Making these annotations is time-consuming. We propose for the first time, an iterative learning and annotation method that is able to detect, segment and annotate instances in datasets composed of multiple similar objects. The approach requires minimal human intervention and needs only a bootstrapping set containing very few annotations. Experiments on two different datasets show the validity of the approach in different applications related to visual inspection. 
\end{abstract}
%

\vspace{-2mm}
\section{Introduction}
\label{sec:intro}
\vspace{-2mm}

Instance segmentation is a well-known computer vision task that involves locating, segmenting, and identifying individual instances of several objects in a set of images. It has multiple application domains, ranging from surveillance to face recognition, but it has special interest in visual inspection tasks such as fault detection or quality monitoring. Classical machine learning algorithms for object detection and instance segmentation are based on, e.g., template matching and keypoint detection. However, they also rely on defining the features of interest beforehand. Recently, deep-learning neural network methods have seen rapid development in this domain, allowing for more complex models that learn the relevant features without designing them ad-hoc for each use case \cite{survey}.

One of the most recent high performing methods is Mask R-CNN \cite{he2018mask}. An extension to Fast R-CNN \cite{girshick2015fast}, it provides an accurate and fast method for detecting and classifying objects in images using region-based convolutional neural networks. Whereas Fast R-CNN only deals with bounding boxes, Mask R-CNN extends the framework to generate segmentation masks inside the detected objects' bounding boxes. Multiple implementations of Mask R-CNN exist\cite{surveymask}. For example, Detectron2 \cite{wu2019detectron2} is a software system that implements state-of-the-art algorithms for object detection, including Mask R-CNN and it can easily be retrained using new datasets for custom instance segmentation tasks.

Training instance segmentation models usually requires high amounts of annotated data. In the usual case, the ground truth labels are generated by hand. However, this is very arduous and time-consuming\cite{humanintheloop}. Very little work exists that leverages semisupervised and iterative learning approaches for annotating datasets using minimum amounts of labelled data, but recent studies \cite{adhikari2020iterative}\cite{sample}\cite{bounding} showed its applicability for automatic bounding box generation. These approaches generates only rough bounding boxes and require human-in-the-loop \cite{humanintheloop} in each iterative round for correcting the annotated boxes and labels, which is not always feasible in visual inspection.

We aimed to minimize the role of manual annotation by implementing a semi-supervised, few shot, self-learning iterative system. In our approach, we build complete annotated datasets with full instance segmentation masks by leveraging minimal initial user input, which is only required in annotating a very small number of class instances only at the start. Subsequent learning of other instances is done by an instance segmentation model that iteratively teaches itself, essentially by generating a new "ground truth" data set on each iteration.

\vspace{-3mm}
\section{Iterative learning system.}
\vspace{-3mm}
The proposed instance segmentation approach consists of three main stages that rely on three partitions of a dataset: The bootstrapping set contains a small number of images with a few user-made annotations of target object instances. The training set contains a larger set of un-annotated images, containing mainly instances of the desired target object. The testing set contains any number of selected images in the whole dataset that are fully annotated for testing purposes) and not present in either the bootstrapping or the training sets. 

\begin{figure}[h]
 \begin{center}
 \vspace{-1mm}
   \includegraphics[width=0.95\linewidth]{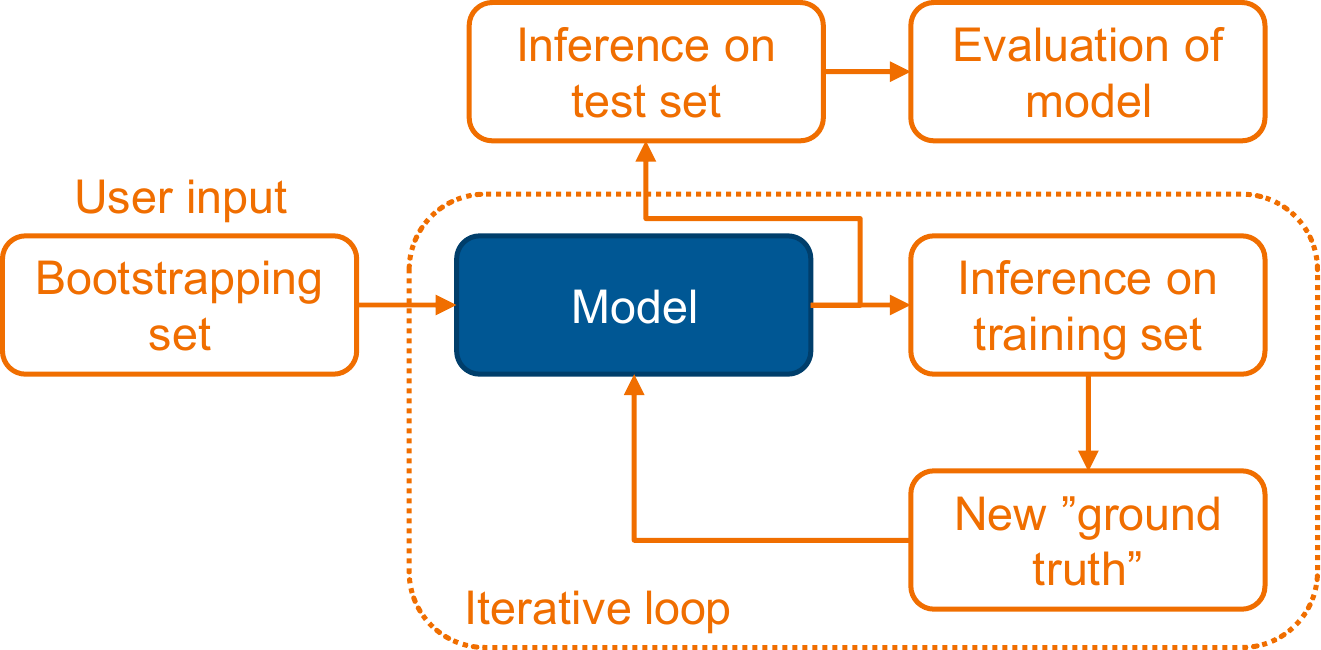}
\end{center}
\vspace{-5mm}
   \caption{Flowchart outlining the developed framework.}
\label{fig:flowchart} 
\vspace{-6mm}
\end{figure}

The three stages involved are the initiation phase, the iterative learning phase and the evaluation phase (see Figure \ref{fig:flowchart}). The initiation phase utilizes the small bootstrapping set for fine tuning an instance segmentation model that has been previously trained on a large multi-class dataset. The result of this phase is the bootstrapping model that will then be used for iterative learning. The iterative learning phase uses the model to run inference on a larger non-annotated training set. The resulting annotations are filtered using the confidence threshold value, where all instance detections over a predetermined threshold are kept and the rest are discarded. These detections are then held as the ground truth, and constitute a new training set that is used in the next iteration round for the training of the model. Subsequent inference results are used again as the new training set until the desired number of iterations is reached. After each training iteration, the model can be evaluated on a external test. 

\vspace{-2mm}
\section{Implementation}
\vspace{-3mm}
We developed the approach using Mask R-CNN in the Detectron2 software system (version 0.1.1) using the R50 FPN 3x instance segmentation baseline from their model zoo. The model has been pretrained on the COCO \cite{coco} 2017 challenge dataset using a Mask R-CNN instance segmentation head. As evaluation metric we use the average precision and recall at confidence level 75\% (AP75 and AR75). As the framework allows for saving the model weights on each iteration, any model can be restored after completing all the iterations.

On each iteration round, the training is continued where the last iteration left off, i.e. the weights of the model are carried and updated throughout the iterative process. Instance segmentation training contains hyperparameters such as the number and type of data augmentations or the batch sizes. In our iterative setup, two additional parameters are the added and become most relevant: the number of epochs in each training iteration round, and the threshold (or confidence) of detection. The threshold controls which instances are carried on to the next iteration; the detected instances below this threshold are dropped since they are considered to be possible missdetections. One epoch consists of inputting a predetermined number of images (in this case 2) in a batch to training a selected number of times  (in this case 24) using different data augmentations each time. The number of epochs determines how many of these batch training rounds are done on each iteration.

\vspace{-5mm}
\section{Methods}
\begin{figure}[h]
    \centering
    \vspace{-2mm}
    \includegraphics[width=0.95\linewidth]{{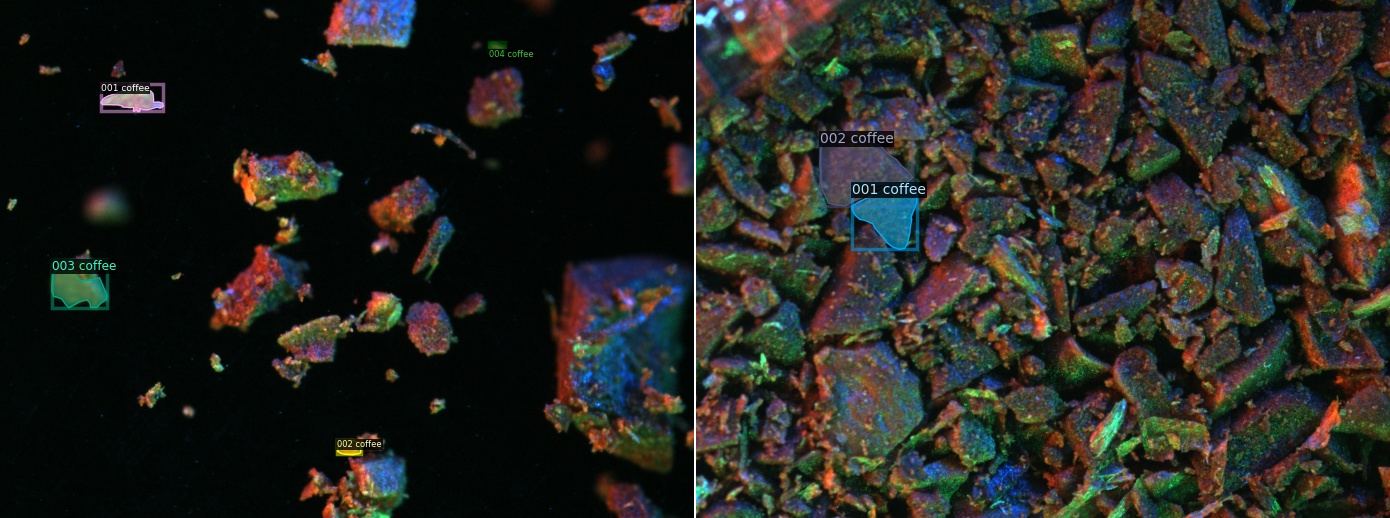}}
    \vspace{-2mm}
    \caption{Bootstrapping images of \textit{coffee} dataset with 6 randomly selected ground truth annotations.}
    \vspace{-2mm}
    \label{fig:train_images}
\end{figure}

\vspace{-3mm}
For testing the system, two different datasets were used. The first dataset \textit{(coffee)}, available in Zenodo \cite{coffee_dataset}, consists of microscope images of coffee ground particles of various irregular shapes and sizes. In some images, the particles are stacked close to one another. Moreover, as the microscope's focus is narrow, the edges of the particles are somewhat blurred in some cases, making it difficult to establish a definite ground truth. The bootstrapping set consists of one image of unconnected coffee grains and one image of connected coffee grains with annotations. The training set includes the bootstrapping set and 50 other images. Finally, the testing set consists of three images, one with unconnected, one with loosely overlapping, and one with heavily connected grains; in total, 252 instances are annotated. Bootstrapping images are shown in Figure \ref{fig:train_images}.

The second dataset \textit{(fruits)} \cite{fruit_dataset} is a public dataset that consists of 250 object instances in 18 photos of 3 different target classes depicting fruits (dates, figs, and hazelnuts) laid on solid surfaces. The target classes are presented together with extra objects with similar shapes wrapped in gold-foil. 

\vspace{-3mm}
\section{Results}
\vspace{-3mm}

We performed extensive experiments on the \textit{(coffee)} dataset. Although it only contains elements of one class, this dataset is very challenging since it contains a high number of objects, with great variability depicted by the various sizes, different focus and distances to the camera, and highly irregular shapes of the particles. We study the performance of our iterative learning approach with respect to the effect of the number of annotations in the bootstrapping set. The experiments are conducted for a total of 15 iterations, and depicted in Table \ref{tab:coffee_annotations}. As it can be seen, even with as low as 1 annotation the model is able to generalize and perform on-par with other runs with more annotations.

\begin{table}[h]
\centering
\vspace{-2mm}
\caption{Results on \textit{coffee} test set inference with constant epoch number 100 and threshold 0.25.}
\label{tab:coffee_annotations}
\begin{tabular}{ |c|c|c|c|c|c| }
 \hline
Number & Best & AP75  & AR75  & Number \\
annotations & iteration & [\%] & [\%] & particles \\
 \hline
1 & 13 & 50.8 & 46.1 & 206 \\
3 & 5 & 42.9 & 32.8 & 209 \\
6 & 15 & 44.0 & 38.9 & 211 \\ 
12 & 5 & 52.8 & 45.4 & 247 \\
18 & 5 & 54.4 & 48.7 & 221 \\
24 & 4 & 56.3 & 53.3 & 206 \\
30 & 8 & \textbf{58.7} & \textbf{53.9} & 209 \\
73 & 6 & 54.8 & 48.4 & 233 \\
\hline
\end{tabular}
\end{table}

Each training iteration needs to be trained for a number of epochs. This parameter has an impact on the performance of the iterative training. The results of our experiments, using a bootstraping set of 6 random annotations, are shown in Table \ref{tab:coffee_epochs}. While a small number of epochs might lead to no instance detections, a number that is too large might result in overfitted models unable to generalize to unseen objects. We show results for different epochs per iteration, conducted for a maximum of 15 iterations with the 6 random annotations seen in Figure \ref{fig:train_images}. 

\begin{table}[h]
\centering
\vspace{-4mm}
\caption{Results on \textit{coffee} test set with constant threshold 0.25 and 6 random annotations.}
\label{tab:coffee_epochs}
\begin{tabular}{ |c|c|c|c|c|c| }
 \hline
Number & Best & AP75  & AR75  & Number of \\
epochs &  iteration & [\%] &  [\%] & instances \\
 \hline
25  & 15 & 44.0 & 38.9 & 211 \\
50  & 7 & 58.4 & 45.2 & 216 \\
100  & 13 & 51.6 & 45.9 & 223 \\
200  & 8 & 40.1 & 32.0 & 202 \\
400  & 13 & 45.2 & 38.6 & 210 \\
\hline
\end{tabular}
\vspace{-2mm}
\end{table}

None of the generated models are able to detect the exact number of instances in the dataset. This was expected due to the complexity of the problem. For the number of detected instances, the number of epochs per iteration has a clear role.  Very small numbers result in no new detections on new images of the training set. On the other hand, with a large number of training epochs per iteration (e.g., 200 or 400), the model seems to overfit to the already seen shapes, and only 30 instances are found in the testing images. This effect is less noticeable with lower thresholds, where lower confidence detections are carried over to the next iteration.

The most important hyperparameter of the system is the threshold selected as the acceptable confidence to carry a particular instance segmentation to the next iteration. To show its effect, we conduct tests for several combinations of thresholds (0.25, 0.50, 0.75), fixing the number to 100 epochs per iteration. The experiments are conducted for a maximum of 50 iterations with the 6 random annotations seen in Figure \ \ref{fig:train_images}, and we depict them in Table \ref{tab:coffee_threshold}.

\begin{table}[h]
\vspace{-4mm}
\centering
\caption{Results on \textit{coffee} test set inference with constant epoch number 100 and different thresholds over 50 iterations.}
\label{tab:coffee_threshold}
\begin{tabular}{ |c|c|c|c|c|c| }
 \hline
Thres- & Best & AP75  & AR75  & Number of \\
hold &  iteration & [\%] &  [\%] & instances \\
 \hline
0.25 & 11 & 52.0 & 45.6 & 234 \\
0.50 & 11 & 44.0 & 39.2 & 152 \\
0.75 & 15 & 22.2 & 20.5 & 74 \\
\hline
\end{tabular}
\vspace{-2mm}
\end{table}

\begin{figure}[h!]
    \centering
    \vspace{-1mm}
    \includegraphics[width=0.95\linewidth]{{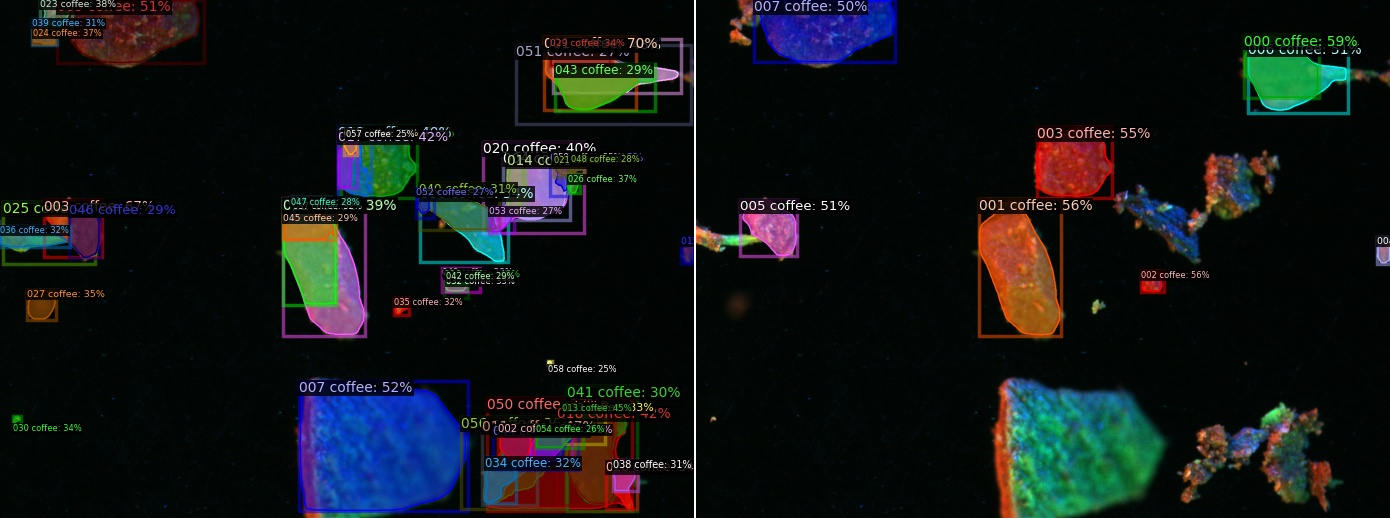}}
    \vspace{-3mm}
    \caption{Segmentations with un-connected particles with threshold 0.25 (left) and 0.50 (right).}
    \label{fig:bootstrapping_inference}
    \vspace{-3mm}
\end{figure}

As expected for a dataset with only one possible target class, the best results for both AP75 and AR75 seem to correspond to a model generated with a low threshold value (0.25), training during 50 epochs per iteration. However, the qualitative analysis of the segmentations seems to suggest that for lower threshold and lower epoch values, the quality of the segmentation might be suboptimal. This can be seen when visually comparing example inferences on the test set for higher threshold values (0.25 vs 0.50), as shown in Figure \ref{fig:bootstrapping_inference}). 

\begin{figure}[h!]
    \centering
    \vspace{-1mm}
    \includegraphics[width=0.95\linewidth]{{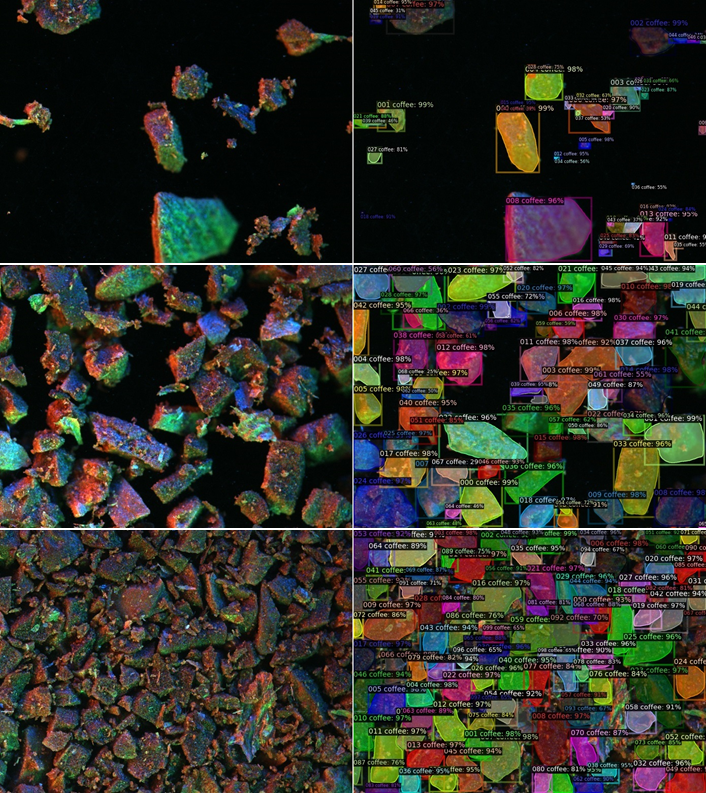}}
    \vspace{-3mm}
    \caption{Three test scenarios (un-connected, loosely overlapping and heavily connected) from the \textit{coffee} dataset (left), and the associated model predictions with threshold 0.25, epoch number 50, iteration round 7 (right).}
    \label{fig:all_both}
    \vspace{-4mm}
\end{figure}

It is especially evident that with lower threshold values, many instances overlap each other regardless of the implemented nonmaximum suppression. We argue that this effect might be due to feature pyramid network scaling inaccuracies \cite{pyramid} that propagate throughout the iterations. Training the model during additional iterations is able to correct some of these inaccuracies. 
The results of the best iteration (number 7) are shown in Figure \ref{fig:all_both}. As can be seen, the segmentations are very accurate for the unconnected and overlapping cases. The heavily connected case is difficult to evaluate exactly, as even the ground truth is very subjective. Nonetheless, many of the particles are detected, spanning the whole image.

In addition, we check the effects posed by the selection of different threshold values in a dataset that contains several target classes, mixed with objects that are not of interest. We conduct the experiments in the \textit{fruit} dataset containing 3 target classes and several undesired objects. For the experiment, we create a minimum bootstrapping set where we annotate just one object of each three classes. We train the models for a fixed number of 10 iterations and 100 epochs for leave-one-image-out validation approach, where one single fully annotated image is held out of the training and is used for testing.  

\begin{figure}[h]
    \centering
    \includegraphics[width=0.85\linewidth]{{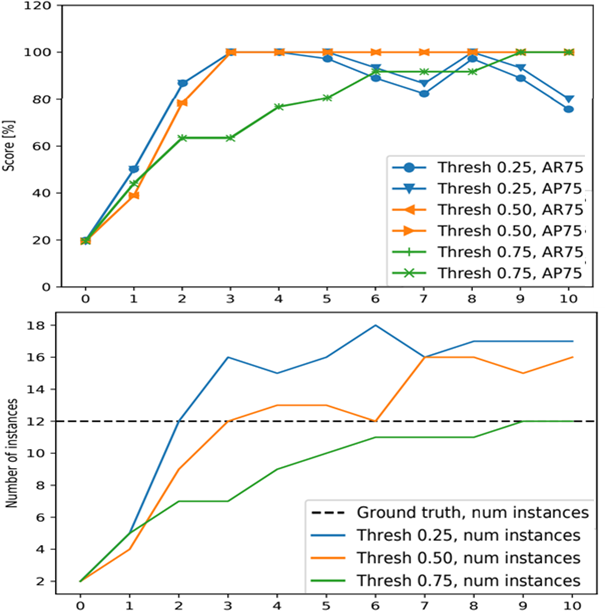}}
    \caption{Results on test set inference of the fruit dataset on each iteration with constant epoch number 100 and varying threshold values (0.25, 0.50, 0.75).}
    \label{fig:threshold_test_plot}
    \vspace{-2mm}
\end{figure}

Figure \ref{fig:threshold_test_plot}, show the results for varying threshold values. The horizontal scale marks the number of iterations: 0 corresponds to training with annotations that belong only to the bootstrapping set, while subsequent iterations include the new annotations found in instances of other images of the training set. The vertical scale on the upper sub-figure is the percentage score of AP75 and AR75. In the lower sub-figure the scale shows the particle number, with the ground truth marked with the black dashed horizontal line.

\begin{figure}[h]
    \centering
    \vspace{-2mm}
    \includegraphics[width=0.92\linewidth]{{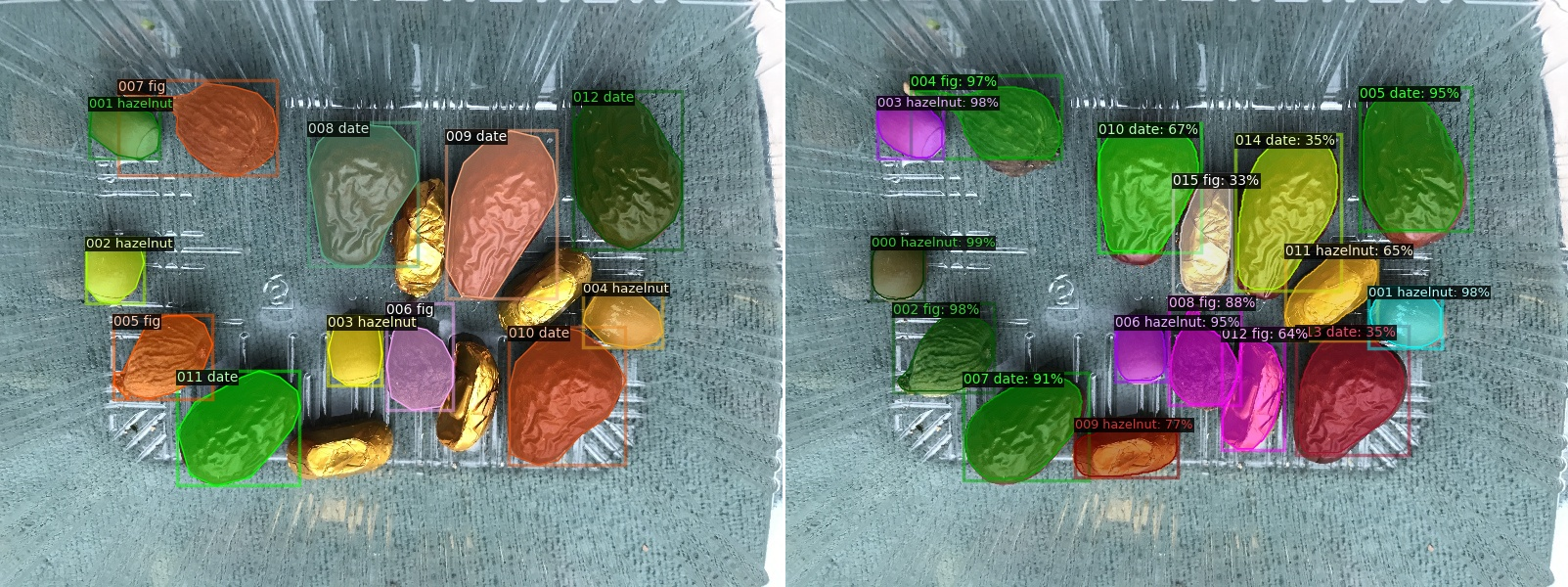}}
    \caption{Segmentations on the test image of the fruit dataset: ground truth (left) and inference on the 3rd iteration for 0.25 threshold and 100 epochs (right).}
    \vspace{-2mm}
    \label{fig:14000_iter_3_and_gt}
    \vspace{-2mm}
\end{figure}

The results show how for a low threshold value of 0.25, the model finds all annotated objects already on the third iteration, as manifested by AP75 value of 100\%. More iterations result in quick generalization to data that does not belong to the desired classes, thus finding more particles in the testing image than the annotated (i.e.\ gold-foiled particles). A qualitative example of this behaviour can be seen in Figure \ref{fig:14000_iter_3_and_gt}.  

Figure \ref{fig:14000_iter_3_and_gt} shows qualitative results of inspecting the best iteration results for a threshold value of 0.25. Although AP75 and AR75 metrics are at 100\%, it can be seen that the segmentations do not seem to encompass the borders of the objects accurately (similarly as with the \textit{coffee} dataset). The results suggest that as we progress through a large number of iterations, the probability of detecting non-target objects becomes more noticeable, even if the threshold is set relatively high. Figure \ref{fig:threshold_test_plot} shows values for a threshold value of 0.5 on the third iteration, that results in the best AP75 and AR75 values and all target objects detected. When increasing the number of iterations, the number of detections fluctuates over the maximum number of target object, due to the model detecting some of the gold-foiled objects in the image set and allocating them to one of the three classes. 

Using a more strict threshold of 0.75, the model slowly converges to find only the target classes and objects. This suggests that for training datasets containing more than one class, more conservative thresholds should be used. Lower thresholds might show faster convergence, but at the cost of more misclassifications and lower quality annotations\cite{threshold}. 

From our experiments, it can be seen that bootstrapping the iterative learning system with  a very low number of annotations is able to converge to high-quality annotations for several cases. However, at the moment it is not possible to select automatically an optimal confidence threshold value and its associated optimal number of epochs per iteration. Common guidelines on the selection can be extracted \cite{strategy}, but after all, these hyperparameters might require fine tuning depending on the particular use case and dataset. 

\vspace{-4mm}
\section{Conclusion}
\vspace{-3mm}
In this paper, a simple and light-weight semisupervised few-shot object detection framework, based on leveraging iterative self-learning, was shown. The framework utilizes an existing state-of-the-art object detection algorithm that is iteratively retrained for customized purposes using minimum amounts of labelled data. The components of the pipeline are readily exchangeable. The system is able to generate high quality segmented and classified datasets from a very small bootstrapping subset of user-annotated instances. The system is useful for training models in multiple visual inspection tasks.

The system is able to produce a well-performing model after a few iterations. However, challenges in the iterative training system still remain. Those are related to non-target objects and shapes present in the images. The model can easily pick up instances outside the training classes due to similarity, which propagate through the self-learning process as the ground truth, further confusing the model. This could be mitigated in the future by including a clustering module that groups similar objects in different classes after each iteration.  

The selection of the best parameters is nontrivial and requires grid searching with potential values, as these values are arguably dependent on the number of target classes and dataset size and quality.

\bibliographystyle{IEEEbib}
\bibliography{iterative_learning}

\end{document}